%% file: main.tex
\documentclass{article} 
\usepackage{arxiv,times}

\input{math_commands.tex}

\usepackage{hyperref}
\usepackage{url}
\usepackage{graphicx} 

\usepackage{booktabs}
\usepackage[table]{xcolor}

\definecolor{groupgray}{RGB}{244,245,247}
\definecolor{skillvblue}{RGB}{232,242,252}
\definecolor{scaleorange}{RGB}{190,105,20}
\usepackage{threeparttable}
\usepackage{tablefootnote}
\newcommand{\best}[1]{\textbf{#1}}
\newcommand{\second}[1]{\underline{#1}}
\usepackage[table]{xcolor}
\usepackage{colortbl}
\usepackage{cleveref}

\usepackage{algorithm}
\usepackage{algpseudocode}
\usepackage{amssymb}

\usepackage[blocks]{authblk} 
\title{Skill-V: Verifiable Self-Evolving Skill Library for Interactive Agents}

\author{
 Jie Ma$^{1,2}$\thanks{This work was completed during an internship at Kuaishou Technology.} ~~ Zhipeng Qian$^{2, 3}$~~ Yufei Ma$^{2}$~~ Zihan Liang$^2$~~ Jiayi Ji$^1$~~ \\  \vspace{-2mm}
 Qingpeng Cai$^2$~~ Ben Chen$^2$~~ Peng Jiang$^2$~~ Xiaoshuai Sun$^{1}$\thanks{Corresponding author.} \\ \vspace{3mm}
 {\small $^1$Xiamen University~~ \small $^2$Kuaishou Technology~~ \small $^3$Nankai University}\\
}

\arxivfinalcopy
\begin{document}

\maketitle

\begin{abstract}
Interactive agents can turn experience into reusable skills, yet existing self-evolving skill libraries primarily improve by accumulating new knowledge. Failures may lead to new skills, while previously stored skills are less often revisited as new evidence arrives. However, growth alone does not ensure reliability, as a retrieved skill may be inapplicable under the current task conditions, and an existing skill may encode a mis-specified operational boundary. Reliable skill evolution therefore requires not only adding knowledge, but also testing and revising what is already stored.
We introduce \textbf{Skill-V}, a verifiable self-evolving skill library. To make stored knowledge testable, we propose representing skills as versioned, falsifiable contracts that link semantic intent to observable behavioral criteria. We use environment outcomes to drive continuous library evolution. Specifically, task failures motivate skill addition, while disagreements between contract evaluations and task outcomes guide revisions to existing skill boundaries. To validate these revisions, we require them to preserve protected semantic constraints and satisfy non-regression criteria for rubric–outcome metrics on historical replay evidence. Finally, we employ an applicability-aware filter to exclude candidates judged confidently inapplicable to the current task.
Across ALFWorld and WebShop, Skill-V achieves success rates of 95.3\% and 85.9\%, respectively, while maintaining a more compact skill library than growth-oriented baselines. Applicability-aware filtering reduces incorrect skill invocations, and outcome-grounded revisions correct mis-specified skill boundaries without degrading performance on previously observed evidence. These results show that reliable skill evolution requires more than accumulating experience: the library must learn which knowledge to retain, when to revise it, and when it should be applied.
\end{abstract}

\input{src/intro}
\input{src/related_work}
\input{src/method}
\input{src/experiment}

\input{src/conclusion}

\bibliography{references}
\bibliographystyle{references}

\appendix
\input{src/appendix}

\end{document}

%% file: math_commands.tex
\usepackage{amsmath,amsfonts,bm}

\def\eqref#1{equation~\ref{#1}}

\def\1{\bm{1}}

\DeclareMathAlphabet{\mathsfit}{\encodingdefault}{\sfdefault}{m}{sl}
\SetMathAlphabet{\mathsfit}{bold}{\encodingdefault}{\sfdefault}{bx}{n}



%% file: src/intro.tex
\section{Introduction}
\label{sec:introduction}

Interactive agents~\citep{openai, gemini, anthropic, deepseek, kimi} are increasingly capable of autonomous learning, a critical step toward recursive self-improvement~\citep{liu2026pathrsi, yin2025godelrsi}. However, retaining that experience in a form that can be reused, inspected, and corrected remains challenging~\citep{ReAct, Reflexion, ExpeL}. Reinforcement learning absorbs feedback into model parameters, making individual strategies difficult to isolate or revise. External skill libraries offer a more explicit alternative. They store reusable procedural knowledge outside the policy, allowing agents to retrieve it to guide future decisions~\citep{Voyager,skillrl}. Recent methods~\citep{skillrl, skill1, Skill-R1, skillos, SAGE_Skill} further allow such libraries to evolve during training by distilling interaction trajectories into new skills. This turns experience into an expanding external memory, but expansion alone does not make that memory reliable.

\begin{figure*}[t]
    \centering
    \includegraphics[width=0.95\textwidth]{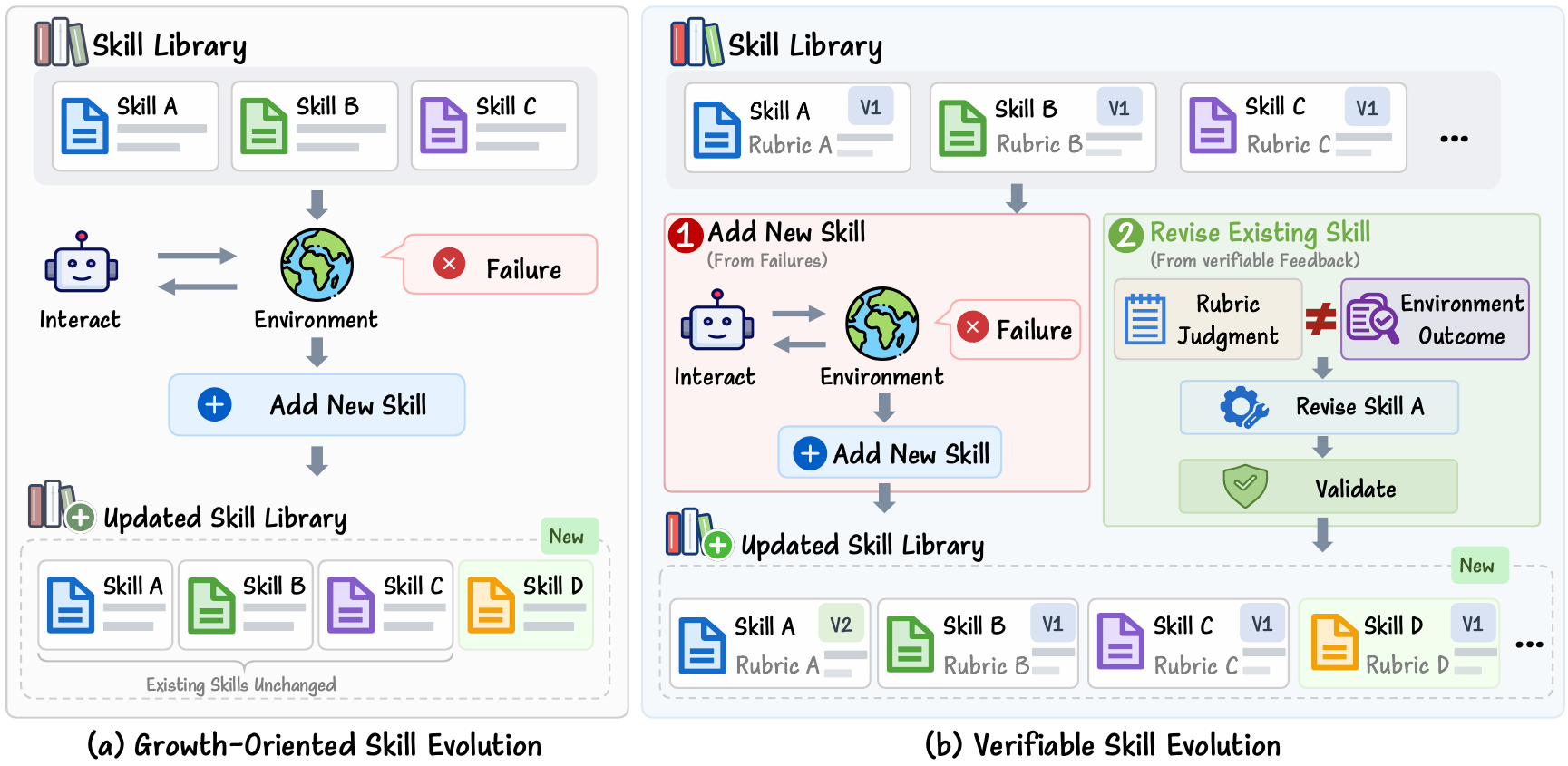}
    \caption{
    \textbf{Comparison of self-evolving skill library paradigms.} 
    \textbf{(a)} Existing methods rely on growth-oriented evolution. Failures prompt the addition of new skills, but previously stored skills are rarely questioned, which often leads to boundary conflicts. 
    \textbf{(b)} \textbf{Skill-V} treats skills as versioned, falsifiable contracts. It adds missing skills from failures (Path 1) and revises existing operational boundaries when rubric judgments disagree with task outcomes (Path 2). Through validated revisions, the library expands its coverage while actively maintaining the reliability of stored knowledge.
        }
    \label{fig:compare}
\end{figure*}

As a skill library grows, the main challenge is no longer only whether it contains enough useful strategies, but whether the strategies already stored remain valid~\citep{skillos, skillops}. Existing self-evolving skill libraries~\citep{skillrl,skill1} primarily improve by accumulating new content. While failures lead to new skills, previously stored skills are rarely revisited (\Cref{fig:compare}(a)). A retrieved skill may be semantically relevant yet inapplicable under the current task conditions~\citep{SkillReranker, skillrouterskillroutingllm, skill1}. A stored skill may also encode an incorrect operational boundary that is either too broad and admits ineffective behavior, or too narrow and excludes a valid strategy. These failures stem not merely from missing knowledge, but from incorrect decisions about when existing knowledge should be trusted. Such unreliability severely limits long-term autonomous self-improvement.

Reliable skill evolution therefore requires more than adding new knowledge~\citep{skillrevise, CRITIC_iclr}. An evolving library must also test whether its existing knowledge remains consistent with new experience and revise it when evidence exposes an incorrect boundary. This motivates our central question:

\vspace{-2.0mm}
\begin{center}
\fcolorbox{gray!40}{gray!5}{%
\begin{minipage}{0.95\linewidth}
\vspace{3pt}
\centering
\textit{How can an agent expand its skill library while continuously testing and correcting the knowledge already stored?}
\vspace{3pt}
\end{minipage}%
}
\end{center}
\vspace{-2.0mm}

To answer this, we propose treating each skill as a versioned, falsifiable contract. In addition to semantic intent and a reusable procedure, a skill must specify its conditions of applicability and the observable behavior expected when it is followed. Interaction outcomes can then serve not only as feedback for learning how to act, but also as empirical evidence for testing what the library currently claims. For instance, a rubric that accepts a failed trajectory is likely too permissive, whereas one that rejects a successful trajectory may be too restrictive.

Based on this idea, we introduce Skill-V, a verifiable self-evolving skill library for interactive agents (\Cref{fig:compare}(b)). Skill-V augments each skill with a contract that links its semantic intent to observable behavioral criteria. During training, each interaction provides two complementary signals for evolution. First, task failures expose missing strategies and drive the addition of new skills. Second, rubric–outcome disagreements identify existing contracts whose boundaries require revision. Candidate revisions are committed only when they preserve the skill's semantic identity and do not regress on previously observed evidence. Skill-V thereby expands its coverage while conservatively improving the reliability of its stored knowledge.

Skill-V also separates semantic relevance from task applicability when using the evolved library. It first retrieves semantically related candidates and then filters skills that are confidently inapplicable to the current task. The selected skills guide the policy throughout the interaction, and the resulting trajectory and outcome provide new evidence for subsequent library evolution. This closes the loop between using stored knowledge and testing whether that knowledge should be retained, expanded, or revised. The library evolves not only by expanding its coverage, but also by making its existing knowledge more selective, correctable, and reliable.

We evaluate Skill-V across embodied household tasks in ALFWorld~\citep{ALFWorld} and online shopping in WebShop~\citep{webshop}. We compare against static and growth-only skill libraries to assess overall task performance. We also analyze the effectiveness of applicability filtering and the ability of counterexample-driven revisions to correct existing skill boundaries.
Empirical results confirm that verifiable evolution yields state-of-the-art capabilities. On WebShop, Skill-V achieves a 92.5 task score and an 85.9\% success rate, outperforming existing memory-augmented agents. On ALFWorld, it achieves a 95.3\% success rate and solves several subtasks perfectly while requiring only a fraction of the library size demanded by accumulation-centric methods. Beyond overall success metrics, our analyses demonstrate that Skill-V effectively corrects boundary errors and mitigates the catastrophic forgetting that typically degrades lifelong learning agents.

Our main contributions are:
\begin{itemize}
    \item We formulate self-evolving skill learning as the joint problem of expanding knowledge coverage and maintaining the reliability of existing skills.
    \item We introduce Skill-V, which combines versioned skill contracts, outcome-grounded skill addition and revision, evidence-preserving validation, and applicability-aware skill use.
    \item We evaluate Skill-V on ALFWorld and WebShop, demonstrating that proactive skill maintenance achieves comparable performance, including success rates of 95.3\% on ALFWorld and 85.9\% on WebShop. Crucially, we show that Skill-V effectively corrects boundary errors and mitigates catastrophic forgetting while avoiding the severe library bloat that plagues accumulation-only baselines.

\end{itemize}

%% file: src/related_work.tex



\section{Related Work}

\paragraph{Memory and skill reuse.}
LLM-based agents can learn from experience by retaining trajectories, reflections, or distilled knowledge~\citep{ReAct, GhostIT, MemGPT}. Reflexion~\citep{Reflexion} and ExpeL~\citep{ExpeL} convert interaction feedback into reusable textual memory, while Voyager\citep{Voyager} organizes experience as executable skills~\citep{Toolformer}. These works establish external memory as a complement to parametric learning. Skill-V focuses on a later-stage problem of how persistent procedural knowledge should be tested and corrected after entering memory.

\paragraph{Self-evolving skill learning.}
Recent methods allow skill libraries to improve through interaction and feedback~\citep{CRITIC_iclr, opro}. SkillRL~\citep{skillrl} recursively expands a hierarchical SkillBank alongside policy learning, while Skill1~\citep{skill1} jointly optimizes skill selection, utilization, and distillation. Skill-R1~\citep{Skill-R1} trains a lightweight generator to iteratively revise instance-level skills for a frozen task model, whereas SkillOS~\citep{skillos} learns a curator that updates an external SkillRepo from delayed downstream feedback. These approaches primarily study skill acquisition, credit assignment, or curation. Skill-V instead asks whether a stored skill remains applicable and behaviorally well-specified as new evidence arrives.

\paragraph{Reliable skill maintenance.}
SkillOpt~\citep{skillopt2026} treats a single skill document as a trainable parameter updated via validation-gated edits. Concurrently, SkillOps~\citep{skillops} represents skills as typed contracts and maintains library-level utility, compatibility, risk, and validation. Skill-V differs in both its setting and learning signal. It studies skill maintenance within interactive reinforcement learning, where the policy and library evolve together. General agent reliability is often approached through post-hoc reflection or self-correction~\citep{Reflexion, CRITIC_iclr}. Skill-V instead uses disagreements between skill rubrics and environment outcomes to systematically identify boundary errors. Candidate revisions are admitted only when they preserve semantic intent and do not regress on observed counterexamples. This strict validation is designed to mitigate the catastrophic forgetting that typically degrades lifelong learning agents~\citep{french1999catastrophic, wang2023continual}. Thus, Skill-V extends skill evolution from acquiring more knowledge to maintaining when existing knowledge should be trusted.

%% file: src/method.tex
\section{Method}
\label{sec:method}

\begin{figure*}[t]
    \centering
    \includegraphics[width=\textwidth]{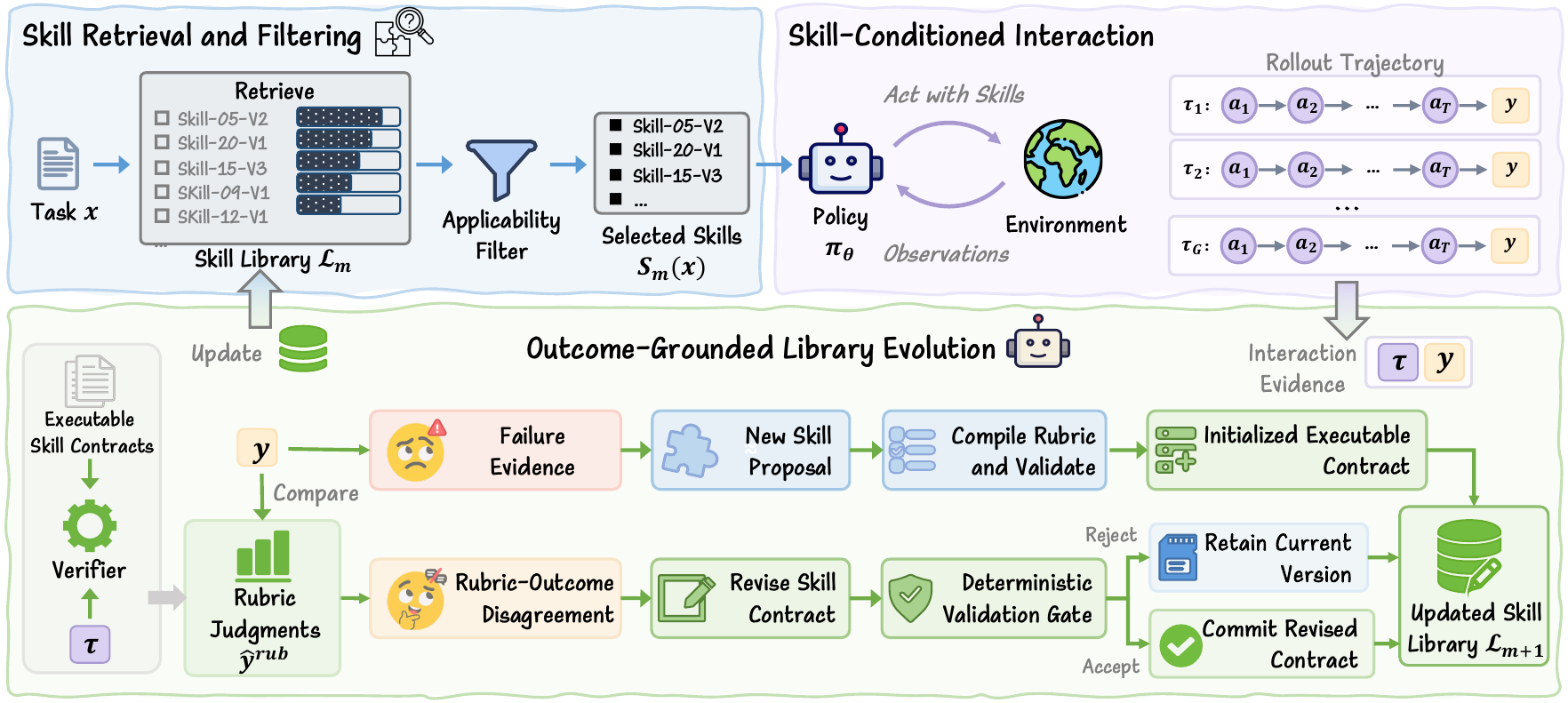}
    \caption{
    \textbf{Overview of Skill-V.}
    Skill-V first retrieves and filters task-relevant skills to guide policy interaction, then uses the resulting trajectories and environment outcomes to evolve the skill library. Failures expose missing strategies and lead to the addition of new semantic skills, whereas rubric--outcome disagreements provide counterexamples for revising existing executable skill contracts. Contract revisions are committed only after deterministic validation, and the updated library is reused in future interactions.
    }
    \label{fig:overview}

\end{figure*}

Skill-V turns an external skill library into an explicit object of learning. As illustrated in~\Cref{fig:overview}, it forms a closed loop between skill use and library evolution. We formalize this interactive setting in~\Cref{sec:problem} and represent skills as versioned, falsifiable contracts in~\Cref{sec:skill_contracts}. During training, environment outcomes drive library evolution through two complementary mechanisms detailed in~\Cref{sec:skill_evolution}. Task failures motivate the addition of new skills, while disagreements between contract rubrics and actual outcomes trigger targeted boundary revisions. Finally, \Cref{sec:skill_use} describes how the agent retrieves and filters these skills to ensure task applicability during inference.

\subsection{Problem Formulation}
\label{sec:problem}

We consider an interactive agent that receives a task instruction \(x\), produces a trajectory \(\tau=(o_1,a_1,\ldots,o_T,a_T)\), and obtains an environment outcome \(y(\tau)\in\{0,1\}\). The agent maintains an external skill library \(\mathcal{L}_m=\{s_i\}_{i=1}^{N_m}\), where \(m\) indexes the current library version. Each skill is a reusable natural-language strategy that describes when a behavior is useful and how it should be carried out. Rather than specifying a complete policy or a fixed action sequence, a skill provides procedural guidance that the policy adapts to the current interaction.

For a task \(x\), the agent retrieves a subset of skills \(\mathcal{S}_m(x)=\operatorname{Retrieve}(x,\mathcal{L}_m)\) (detailed in \Cref{sec:skill_use}) and conditions the policy on the selected skills:
\begin{equation}
a_t
\sim
\pi_\theta
\left(
a_t
\mid
x,h_t,\mathcal{S}_m(x)
\right),
\label{eq:skill_conditioned_policy}
\end{equation}
where \(h_t=(o_1,a_1,\ldots,o_t)\) denotes the interaction history.

During training, the agent collects experience \(\mathcal{D}_m=\{(x_n,\tau_n,y_n)\}_{n=1}^{M}\) and updates the library as \(\mathcal{L}_{m+1}=\operatorname{Update}(\mathcal{L}_m,\mathcal{D}_m)\). The update may add a missing skill, revise an existing skill, or leave the library unchanged. The central problem is therefore not simply how to grow the library, but how to determine which change is supported by experience.

\subsection{Skills as Verifiable Contracts}
\label{sec:skill_contracts}

Skill-V represents each skill \(s_i\in\mathcal{L}_m\) as
\begin{equation}
s_i =
\left(
\phi_i,\,
p_i,\,
\rho_i,\,
\nu_i
\right),
\qquad
\rho_i\in
\left\{
\emptyset,\,
(A_i,\mathcal{C}_i)
\right\},
\label{eq:skill_representation}
\end{equation}
where \(\phi_i\) specifies the semantic intent and intended applicability of the skill, \(p_i\) describes its reusable procedure, and \(\nu_i\) records its skill and rubric versions. The rubric \(\rho_i\) remains \(\emptyset\) while its executable specification is pending. Otherwise, it comprises applicability conditions \(A_i\) and a set of weighted behavioral criteria \(\mathcal{C}_i=\{c_{ij}\}_j\).

The semantic specification determines what knowledge the skill represents and how it should guide the policy. However, free-form language alone does not provide a systematic way to test whether the skill remains consistent with experience. The executable rubric therefore captures observable implications of the skill, including required steps, forbidden behaviors, preconditions, postconditions, and termination conditions. Each criterion is grounded in evidence from the task instruction, executed actions, observations, or event sequences. Some criteria are designated as hard constraints and must be satisfied regardless of the aggregate rubric score. Ultimately, these two components serve complementary roles. The semantic specification defines what the skill means, while the executable rubric makes its current operational boundary falsifiable. Skill-V can therefore revise when and how a skill should apply without silently replacing the knowledge that the skill is intended to represent.

\paragraph{Contract initialization.}
Existing and newly added skills may initially contain only semantic descriptions and reusable procedures, resulting in \(\rho_i=\emptyset\). Such skills can already guide the policy but cannot yet participate in executable contract evaluation. Given applicable experience, a contract updater proposes observable criteria based on the interaction history. Provided this proposal preserves the skill's semantic intent, admits non-empty executable support, and satisfies the contract specification, it becomes the first executable version of the skill. Subsequent modifications are recorded as new versions. This strict versioning makes the evolution of each skill explicit and traceable.

\subsection{Outcome-Grounded Skill Evolution}
\label{sec:skill_evolution}

Skill-V uses trajectories and environment outcomes as interaction evidence \(e=(x,\tau,y)\) to derive two complementary signals for skill library evolution. A task failure suggests that the library may lack a reusable strategy, motivating the addition of a new skill.  Furthermore, to identify existing contracts that may be mis-specified, Skill-V evaluates the executable contracts that apply to task \(x\). Let \(\mathcal{A}(x)=\{s_i\in\mathcal{L}_m:A_i(x)=1\}\) denote this set of applicable skills. For each criterion \(c_{ij}\) within these contracts, \(m_{ij}(\tau)\in\{0,1\}\) indicates whether the trajectory satisfies the criterion, and \(w_{ij}\) denotes its assigned weight. Skill-V aggregates these applicable criteria to produce an overall rubric judgment:
\begin{equation}
q(\tau) = \frac{\sum_{s_i\in\mathcal{A}(x)} \sum_j w_{ij}m_{ij}(\tau)}{\sum_{s_i\in\mathcal{A}(x)} \sum_j w_{ij}}.
\label{eq:rubric_score}
\end{equation}
The final contract-level judgment is then produced by applying a strict threshold and checking for hard constraint violations:
\begin{equation}
\hat{y}^{\,\mathrm{rub}}(\tau) = \mathbb{I} \left[ q(\tau)\geq\eta \ \land\ H(\tau)=0 \right],
\label{eq:rubric_decision}
\end{equation}
where \(H(\tau)\) indicates whether any applicable hard criterion is violated. When no executable criterion applies, Skill-V does not produce a contract-level judgment for that trajectory.

Alongside the aggregate judgment, Skill-V retains individual rubric scores and failed criteria to identify applicable contracts that contradict the trajectory. 
Comparing rubric judgments with environment outcomes provides diagnostic evidence for skill maintenance. A rubric pass followed by task failure may indicate an overly permissive or incomplete executable specification, whereas a rubric failure on a successful trajectory may indicate an overly restrictive criterion. Agreement cases provide supporting or negative evidence, and joint failures may also motivate procedural repair. Because environment outcomes are defined for complete trajectories, these signals do not establish that a particular skill caused success or failure. Skill-V uses them to prioritize candidate updates, which are subsequently screened by the validation gate.

\paragraph{Skill addition.}
A failed trajectory may reveal that the current library lacks an appropriate strategy. To address this capability gap, a language-model updater analyzes the failed interaction to propose new procedural knowledge. The system adds this resulting skill in semantic form with its executable rubric initially pending. The skill nevertheless becomes immediately available for subsequent interactions. This proactive addition continuously expands the library's coverage.

\paragraph{Skill revision and validation.}
A disagreement between the rubric judgment and environment outcome provides a counterexample to the current operational boundary. Using the per-skill diagnostics, the updater proposes targeted revisions to the relevant contracts. A proposal may refine the skill's applicability conditions, procedure, or executable criteria, but must preserve the protected semantic constraints that encode the identity of the skill. However, such counterexamples do not justify arbitrary rewrites. For each skill \(s_i\), let \(\mathcal{E}_i\) denote its observed evidence set, which is maintained via a bounded replay buffer of applicable trajectories. Skill-V evaluates the current rubric \(\rho_i\) and candidate rubric \(\rho_i'\) by treating rubric pass/fail as a prediction of the environment outcome. We measure their consistency using the overall disagreement rate \(\operatorname{Disc}\), the false-negative rate on successful trajectories \(\operatorname{FNR}\), and balanced accuracy \(\operatorname{BAcc}\).

We require revisions to be non-regressive on observed evidence. Specifically, a candidate \(\rho_i'\) is considered evidence-preserving relative to \(\rho_i\) only if it yields a lower or equal \(\operatorname{Disc}\) and \(\operatorname{FNR}\), along with an equal or higher \(\operatorname{BAcc}\) on the evidence set \(\mathcal{E}_i\). To be eligible for acceptance, this evidence-preserving revision must also maintain non-empty executable support, satisfy the contract specification, and preserve the semantic constraints of its source skill. Furthermore, if the revision alters a hard executable criterion, Skill-V requires a strict improvement in at least one of these three statistical metrics. An accepted revision is committed as the next version of the skill contract, whereas failing candidates are rejected to retain the current version. This rigorous constraint allows Skill-V to refine a falsified boundary without sacrificing behavior that remains supported by experience.

\subsection{Applicability-Aware Skill Use}
\label{sec:skill_use}

The evolved skill library must determine which skills are useful for a particular task. Semantic similarity provides an efficient first-stage retrieval signal, but a semantically related skill may not be applicable under the current task constraints. Skill-V therefore separates candidate retrieval from applicability judgment. Given an embedding function \(f\), Skill-V ranks stored skills by the cosine similarity between the task embedding \(f(x)\) and their semantic representations. It retrieves an expanded candidate set from both general and task-specific portions of the library, allowing the applicability stage to inspect more candidates than will eventually be provided to the policy.

To evaluate task applicability, Skill-V combines deterministic contract checks with an LLM-based applicability judge. The judge inspects each candidate with respect to the current task and returns one of \(\{\textsc{Applicable}, \textsc{NotApplicable}, \textsc{Uncertain}\}\) together with a confidence score.
Skill-V removes a candidate only when it is judged not applicable with confidence above a threshold \(\delta\). This conservative rule retains uncertain but potentially useful skills while filtering confident mismatches. Among the remaining candidates, Skill-V retains those whose semantic similarity lies within a margin \(\epsilon\) of the highest-scoring candidate and selects the top \(K\) skills. If applicability judgments are unavailable, Skill-V falls back to semantic Top-\(K\) retrieval. The selected skills are provided to the policy at the beginning of the episode and used throughout the interaction. This execution closes the learning loop. The skill library guides the policy, the environment evaluates the resulting behavior, and the observed outcome provides new interaction evidence for adding missing skills or revising mis-specified contracts.

%% file: src/experiment.tex
\section{Experiments}
\label{sec:experiments}

\subsection{Experimental Setup}
\label{sec:experimental_setup}

\paragraph{Environments and metrics.}
We evaluate Skill-V on ALFWorld~\citep{ALFWorld} and WebShop~\citep{webshop} using the same evaluation protocol as SkillRL and Skill1~\citep{skillrl,skill1}. ALFWorld is a text-based embodied environment requiring agents to complete multi-step tasks. We report the overall success rate alongside success rates across its six task types (Pick, Look, Clean, Heat, Cool, and Pick2). WebShop evaluates interactive search and purchase under natural-language constraints, where we report both the average task score and success rate.

\paragraph{Baselines.}
We compare Skill-V against a comprehensive suite of baselines, including closed-source LLM agents, prompting- and memory-based methods, RL-based agents, and memory-augmented RL methods. This latter category serves as our primary point of comparison and includes MemRL~\citep{MemRL}, EvolveR~\citep{EvolveR}, Mem0+GRPO~\citep{Mem0}, SimpleMem+GRPO~\citep{SimpleMem}, SkillRL~\citep{skillrl}, RetroAgent~\citep{RetroAgent}, and Skill1~\citep{skill1}.

\paragraph{Implementation details.}
Skill retrieval uses Qwen3-Embedding-0.6B, followed by an applicability judge
implemented with DeepSeek-V4-Flash~\citep{deepseek}. We retrieve up to six general skills and five task-specific skills after filtering. The initial candidate pool is twice the final retrieval budget, and candidates judged inapplicable with confidence above \(0.8\) are discarded. Executable rubrics employ a pass threshold of \(\eta=0.7\), and the skill library is periodically evolved using recent training trajectories. Complete details and hyperparameters are provided in Appendix~\ref{app:implementation} and \ref{app:experimental_details}.


\begin{table*}[t]
\centering
\caption{
Main results on ALFWorld and WebShop.
ALFWorld reports success rates (\%) for each task type and overall.
WebShop reports the average task score and success rate (\%).
The best result is shown in \textbf{bold} and the second best is \underline{underlined}.
\textsuperscript{\textcolor{scaleorange}{\(\dagger\)}}Skill1 expands its ALFWorld skill library to its capacity of \(5{,}000\) skills; stored units may not be directly comparable across methods.}
\label{tab:main_results}

\begingroup
\small
\renewcommand{\arraystretch}{1.08}
\setlength{\tabcolsep}{3.5pt}

\begin{tabular*}{\textwidth}{
@{\extracolsep{\fill}}
lccccccccc
@{}
}
\toprule
& \multicolumn{7}{c}{\textbf{ALFWorld}} 
& \multicolumn{2}{c}{\textbf{WebShop}} \\
\cmidrule(lr){2-8}
\cmidrule(lr){9-10}
\textbf{Method}
& Pick & Look & Clean & Heat & Cool & Pick2
& \textbf{All}
& \textbf{Score} & \textbf{Succ.} \\
\midrule

\rowcolor{groupgray}
\multicolumn{10}{l}{\textit{Closed-source LLMs}} \\
GPT-4o
& 75.3 & 60.8 & 31.2 & 56.7 & 21.6 & 49.8 & 48.0
& 31.8 & 23.7 \\
Gemini-2.5-Pro
& 92.8 & 63.3 & 62.1 & 69.0 & 26.6 & 58.7 & 60.3
& 42.5 & 35.9 \\

\addlinespace[2pt]
\rowcolor{groupgray}
\multicolumn{10}{l}{
\textit{Training-free agents with Qwen2.5-7B-Instruct}
} \\
Qwen2.5
& 33.4 & 21.6 & 19.3 & 6.9 & 2.8 & 3.2 & 14.8
& 26.4 & 7.8 \\
ReAct
& 48.5 & 35.4 & 34.3 & 13.2 & 18.2 & 17.6 & 31.2
& 46.2 & 19.5 \\
Reflexion
& 62.0 & 41.6 & 44.9 & 30.9 & 36.3 & 23.8 & 42.7
& 58.1 & 28.8 \\
Mem0
& 54.0 & 55.0 & 26.9 & 36.4 & 20.8 & 7.7 & 33.6
& 23.9 & 2.0 \\
ExpeL
& 21.0 & 67.0 & 55.0 & 52.0 & 71.0 & 6.0 & 46.3
& 30.9 & 11.2 \\
MemP
& 54.3 & 38.5 & 48.1 & 56.2 & 32.0 & 16.7 & 41.4
& 25.3 & 6.4 \\
SimpleMem
& 64.5 & 33.3 & 20.0 & 12.5 & 33.3 & 3.84 & 29.7
& 33.2 & 8.59 \\

\addlinespace[2pt]
\rowcolor{groupgray}
\multicolumn{10}{l}{
\textit{RL-trained agents with Qwen2.5-7B-Instruct}
} \\
RLOO
& 87.6 & 78.2 & 87.3 & 81.3 & 71.9 & 48.9 & 75.5
& 80.3 & 65.7 \\
GRPO
& 90.8 & 66.1 & 89.3 & 74.7 & 72.5 & 64.7 & 77.6
& 79.3 & 66.1 \\

\addlinespace[2pt]
\rowcolor{groupgray}
\multicolumn{10}{l}{
\textit{Memory-augmented RL agents with Qwen2.5-7B-Instruct}
} \\
MemRL
& 62.8 & 38.5 & 22.2 & 12.5 & 8.0 & 0.0 & 21.4
& 29.5 & 9.2 \\
EvolveR
& 64.9 & 33.3 & 46.4 & 13.3 & 33.3 & 33.3 & 43.8
& 42.5 & 17.6 \\
Mem0+GRPO
& 78.1 & 54.8 & 56.1 & 31.0 & 65.0 & 26.9 & 54.7
& 58.1 & 37.5 \\
SimpleMem+GRPO
& 89.5 & 36.3 & 60.0 & 50.0 & 64.9 & 26.3 & 62.5
& 67.8 & 46.9 \\
SkillRL
& \second{97.9} & 71.4 & 90.0 & 90.0 & \second{95.5}
& 87.5 & 89.9 & 85.2 & 72.7 \\
RetroAgent
& \second{97.9} & 90.9 & \second{99.2} & 92.9 & 85.3
& 91.0 & 94.9 & 88.9 & 82.3 \\
Skill1\textsuperscript{\textcolor{scaleorange}{\(\dagger\)}}
& \best{100.0} & \second{98.6} & 97.3 & \second{99.2}
& \best{96.1} & \best{96.5} & \best{97.5}
& \second{89.7} & \second{82.9} \\

\addlinespace[1pt]
\specialrule{0.7pt}{1pt}{0pt}
\rowcolor{skillvblue}
\textbf{Skill-V (Ours)}
& 93.3
& \best{100.0}
& \best{100.0}
& \best{100.0}
& 90.9
& \second{92.9}
& \second{95.3}
& \best{92.5}
& \best{85.9} \\
\bottomrule
\end{tabular*}
\endgroup
\vspace{-1em}
\end{table*}

\subsection{Main Results}
\label{sec:main_results}

\paragraph{Skill-V substantially improves over task-reward RL.}
Table~\ref{tab:main_results} demonstrates the strong performance of Skill-V across both interactive domains. Compared with GRPO, Skill-V improves the overall ALFWorld success rate from \(77.6\%\) to \(95.3\%\), a gain of \(17.7\) percentage points. On WebShop, it improves the average score from \(79.3\) to \(92.5\) and the success rate from \(66.1\%\) to \(85.9\%\). Since both methods optimize the policy using environment task rewards, these gains suggest that evolving reusable external knowledge provides benefits beyond pure policy optimization.

\paragraph{Skill-V is competitive with the strongest skill-augmented agents.}
As the most direct baseline, SkillRL also combines reinforcement learning with an evolving skill library. Compared to SkillRL, Skill-V elevates the overall ALFWorld success rate from \(89.9\%\) to \(95.3\%\). The improvement spans several task types, including gains of \(28.6\), \(10.0\), \(10.0\), and \(5.4\) percentage points on Look, Clean, Heat, and Pick2, respectively. Skill-V consequently reaches \(100\%\) success on Look, Clean, and Heat. The gains are even larger on WebShop, where Skill-V improves the average score from \(85.2\) to \(92.5\) and the success rate from \(72.7\%\) to \(85.9\%\). These results indicate that verifying skill applicability and revising existing skill boundaries provide substantial benefits beyond acquisition-based skill evolution. 
Skill-V demonstrates similar superiority over more recent memory-augmented agents. It outperforms RetroAgent by \(0.4\) percentage points on ALFWorld, and achieves gains of \(3.6\) points in both average score and success rate on WebShop. While Skill-V marginally trails Skill1 by \(2.2\) percentage points on the overall ALFWorld metric, it achieves notable gains of \(2.8\) points in average score and \(3.0\) percentage points in success rate on WebShop.

\paragraph{Competitive performance does not require aggressive library growth.}
The comparison with Skill1 reveals a substantial difference in library scale. Skill1 expands its ALFWorld library to its capacity of \(5{,}000\) skills~\citep{skill1}, whereas Skill-V contains approximately \(200\) stored entries at the end of training. This represents less than \(4\%\) of the library size demanded by Skill1. Nevertheless, Skill-V achieves \(100\%\) success on the Look, Clean, and Heat task types, outperforming Skill1 by \(1.4\), \(2.7\), and \(0.8\) percentage points, respectively. The combination of a highly compact library and superior subtask performance proves that effective skill evolution depends not only on how much knowledge is accumulated, but also on whether stored skills are strictly applicable and correctly specified.
By prioritizing verifiable maintenance over unrestrained growth, Skill-V successfully circumvents the severe knowledge bloat that typically degrades long-term agent autonomy.

\subsection{Ablation Studies}
\label{sec:ablation}

\paragraph{Validated skill evolution provides gains beyond policy learning.}
Because the policy and skill library evolve together during training, final task performance alone cannot reveal whether the improvement is encoded in the policy parameters or maintained in the external library. We disentangle these sources by fixing the policy checkpoint at the end of training and varying only the external knowledge provided during evaluation (Table~\ref{tab:knowledge_source}).
Without external skills, the frozen policy achieves a \(78.59\%\) success rate. Providing the initial skill library raises performance to \(88.99\%\), demonstrating that external procedural knowledge remains highly useful even after extensive policy training. Using the fully evolved Skill-V library further improves success to \(95.30\%\). Unlike a library expanded through acquisition alone, this library contains newly acquired skills alongside existing contracts revised through rubric-outcome diagnosis and non-regressive validation. This additional 6.31-point gain over the initial library confirms that the evolved skill library provides useful knowledge beyond the initial library, even when the policy parameters are fixed.

\begin{table}[!t]
    \centering
    \small
    \renewcommand{\arraystretch}{1.15} 
    
    \begin{minipage}[t]{0.44\linewidth}
        \centering
        \caption{
        Knowledge-source analysis on ALFWorld. We fix the policy at the end of training and vary only the external knowledge available during evaluation.
        }
        \label{tab:knowledge_source}
        \setlength{\tabcolsep}{2pt} 
        \renewcommand{\arraystretch}{1.4} %
        \begin{tabular}{lccc}
        \toprule
        \textbf{\shortstack{Knowledge \\ Condition}} 
        & \textbf{Success} 
        & \textbf{\shortstack{\(\Delta\) vs.\\No Skill}} 
        & \textbf{\shortstack{\(\Delta\) vs.\\Initial}} \\
        \midrule
        No External Skills    & 78.59 & -- & -- \\
        Initial Skill Library & 88.99 & +10.40 & -- \\
        \textbf{Skill-V Library} & \textbf{95.30} & \textbf{+16.71} & \textbf{+6.31} \\
        \bottomrule
        \end{tabular}
    \end{minipage}\hfill
    \begin{minipage}[t]{0.52\linewidth}
        \centering
        \caption{
        Ablation of skill evolution mechanisms. Failure-based revision corrects permissive contracts, whereas success-based revision corrects restrictive ones.
        }
        \label{tab:evolution-ablation}
        \setlength{\tabcolsep}{4pt}
        \begin{tabular}{lcccc}
        \toprule
        \textbf{Method} 
        & \textbf{Acquire} 
        & \textbf{\shortstack{Fail\\Rev.}} 
        & \textbf{\shortstack{Succ\\Rev.}} 
        & \textbf{\shortstack{Success\\(\%) \(\uparrow\)}} \\
        \midrule
        Static Library   & \(\times\) & \(\times\) & \(\times\) & 73.40 \\
        Acquisition Only & \(\checkmark\) & \(\times\) & \(\times\) & 68.80 \\
        Failure-only     & \(\checkmark\) & \(\checkmark\) & \(\times\) & 67.20 \\
        \textbf{Full Skill-V} & \(\checkmark\) & \(\checkmark\) & \(\checkmark\) & \textbf{95.30} \\
        \bottomrule
        \end{tabular}
    \end{minipage}
    \vspace{-1em}
\end{table}

\paragraph{Reliable evolution requires more than skill accumulation.}
Adding new skills alone fails to recover the performance of the full method (Table~\ref{tab:evolution-ablation}). Specifically, the Acquisition Only variant reaches \(68.8\%\) success, compared with \(95.3\%\) for Skill-V. Restricting revision exclusively to failed trajectories is also insufficient, reaching only \(67.2\%\). In contrast, full Skill-V combines skill acquisition with both directions of contract correction. Failed trajectories expose missing strategies and overly permissive boundaries, while successful counterexamples identify overly restrictive boundaries. Together, these mechanisms outperform all incomplete variants by at least \(21.9\) points.
These results strongly support our central claim that reliable skill evolution benefits from combining library expansion with bidirectional boundary correction.

\begin{table*}[t]
\centering
\small
\caption{
Effect of the Evidence Gate on ALFWorld task performance and revision reliability. Non-regressive revisions do not worsen Disc, FNR, or BAcc on replay evidence. Passing the full gate additionally requires satisfying all remaining acceptance conditions. Regression counts are not mutually exclusive.
}
\label{tab:evidence-gate}
\begin{tabular}{l c c cc ccc}
\toprule
& \textbf{Task perf.}
& \textbf{Revisions}
& \multicolumn{2}{c}{\textbf{Retrospective audit}}
& \multicolumn{3}{c}{\textbf{Metric regressions $\downarrow$}} \\
\cmidrule(lr){2-2}
\cmidrule(lr){3-3}
\cmidrule(lr){4-5}
\cmidrule(lr){6-8}
\textbf{Variant}
& \textbf{\shortstack{Success\\(\%) $\uparrow$}}
& \textbf{Committed}
& \textbf{\shortstack{Non-\\regressive}}
& \textbf{\shortstack{Pass full\\gate}}
& \textbf{Disc}
& \textbf{FNR}
& \textbf{BAcc} \\
\midrule
w/o Evidence Gate
& 75.00
& 46
& 37 (80.40\%)
& 27 (58.70\%)
& 9
& 0
& 3 \\
\textbf{Full Skill-V}
& \textbf{95.30}
& 10
& \textbf{10 (100.00\%)}
& \textbf{10 (100.00\%)}
& \textbf{0}
& \textbf{0}
& \textbf{0} \\
\bottomrule
\end{tabular}
\vspace{-1em}
\end{table*}

\paragraph{Evidence Gate filters regressive skill revisions.}
Structural validity alone does not guarantee that a revision preserves contract reliability. Without evidence-based validation, \(9\) of the \(46\) committed revisions regress on at least one replay metric, and only \(27\) satisfy the complete acceptance criteria under retrospective evaluation. In contrast, all revisions committed by full Skill-V pass the complete gate. Consequently, enforcing this validation improves the overall task success rate from \(75.0\%\) to \(95.3\%\). This substantial gain provides consistent downstream evidence that preventing unsupported revisions critically impacts library evolution.

\begin{table}[!t]
    \centering
    \small
    
    \begin{minipage}[t]{0.54\linewidth}
        \centering
        \caption{
        Effect of the LLM applicability judge on ALFWorld. Isolates the contribution of LLM-based applicability judgment.
        }
        \label{tab:applicability-ablation}
        \setlength{\tabcolsep}{3pt} 
        \begin{tabular}{lcccc}
        \toprule
        \textbf{Variant} 
        & \textbf{\shortstack{Embed.\\Retr.}} 
        & \textbf{\shortstack{Contract\\Check}} 
        & \textbf{\shortstack{LLM\\Judge}} 
        & \textbf{\shortstack{Success\\(\%) \(\uparrow\)}} \\
        \midrule
        Semantic Retr. Only & \(\checkmark\) & \(\times\) & \(\times\) & 71.90 \\
        w/o LLM Judge         & \(\checkmark\) & \(\checkmark\) & \(\times\) & 89.10 \\
        \textbf{Full Skill-V} & \(\checkmark\) & \(\checkmark\) & \(\checkmark\) & \textbf{95.30} \\
        \bottomrule
        \end{tabular}
    \end{minipage}\hfill
    \begin{minipage}[t]{0.42\linewidth}
        \centering
        \caption{
        Effect of rubric-based process reward. The default policy uses only the environment task reward.
        }
        \label{tab:process_reward}
        \setlength{\tabcolsep}{5pt}
        \begin{tabular}{lcc}
        \toprule
        \textbf{Method} 
        & \textbf{\shortstack{Success\\(\%) \(\uparrow\)}} 
        & \textbf{\shortstack{\(\Delta\) vs.\\Default}} \\
        \midrule
        \textbf{Skill-V (default)} & \textbf{95.30} & -- \\
        + Process reward           & 93.80 & \(-1.5\) \\
        \bottomrule
        \end{tabular}
    \end{minipage}
    \vspace{-1em}
\end{table}



\paragraph{Applicability-aware routing improves skill use.}
Relying solely on semantic retrieval severely limits the success rate to \(71.9\%\) (Table~\ref{tab:applicability-ablation}). Retaining deterministic contract checks recovers performance to \(89.1\%\), omitting the LLM judge still drops success from the full \(95.3\%\). Throughout training, this judge labels \(29.6\%\) of retrieved candidates as confidently inapplicable, confirming that semantic similarity frequently surfaces irrelevant skills that misguide the policy. These stepped ablations separate the critical contributions of both deterministic and contextual filtering.

\paragraph{Skill-V performs strongly without additional reward shaping.}
Although Skill-V uses rubrics to diagnose and revise skill contracts, these rubrics could theoretically be converted into process rewards to directly supervise the policy. This raises the alternative hypothesis that the benefits of Skill-V stem from reward shaping rather than improved external knowledge. We examine this possibility by augmenting the default Skill-V objective with a rubric-based process reward (Table~\ref{tab:process_reward}). Incorporating this dense reward yields a slightly lower success rate of \(93.8\%\), compared to \(95.3\%\) for the default configuration. Consequently, we retain the simpler task-reward-only objective for our main experiments. Although process rewards might prove beneficial in other contexts, this comparison confirms that Skill-V achieves its strong performance through verifiable knowledge evolution rather than dense reward shaping.


%% file: src/conclusion.tex
\section{Conclusion}
\label{sec:conclusion}

We introduced Skill-V, a verifiable self-evolving skill library that shifts agent learning from pure knowledge accumulation to rigorous knowledge maintenance. By representing skills as falsifiable contracts, our framework uses task failures to drive skill addition and leverages contract-environment disagreements to trigger targeted boundary revisions. 
An evidence gate ensures these self-generated updates remain entirely non-regressive. 
During inference, an applicability-aware filter effectively excludes confidently inapplicable candidates, preventing irrelevant skills from misguiding the policy. Empirical evaluations across ALFWorld and WebShop demonstrate that Skill-V achieves state-of-the-art performance while maintaining a highly compact library, successfully circumventing the catastrophic forgetting that plagues growth-oriented methods. As interactive agents advance toward recursive self-improvement, the ability to safely verify and correct procedural memory becomes critical. By prioritizing continuous maintenance over unrestrained expansion, Skill-V provides a principled foundation for reliable and autonomously improving agents.

%% file: src/appendix.tex

\appendix

\definecolor{skillvheader}{HTML}{D9E2F3}
\definecolor{skillvsemantic}{HTML}{EAF3FB}
\definecolor{skillvrubric}{HTML}{EAF6EC}
\definecolor{skillvevolution}{HTML}{FFF2E2}

\section{Limitations and Broader Impacts}
\label{app:limitations}

\paragraph{Environment types.}
Skill-V evaluates contracts using discrete text-based criteria. Applying this framework to visual or continuous control environments will require new multimodal evaluation mechanisms. Future research could integrate vision-language models to visually verify physical state changes. Alternatively, continuous environments might benefit from programmatic reward functions acting as executable rubrics.

\paragraph{Skill deletion.}
Our framework only updates operational boundaries to prevent unintended behavior changes. If an agent initially acquires a fundamentally flawed strategy, the system will tightly restrict its use rather than discarding it entirely. Removing these inefficient strategies will therefore require additional mechanisms for explicit skill deletion. Future systems could implement automated skill deprecation to permanently retire skills that consistently fail applicability checks.

\paragraph{Historical dependency.}
The strict validation step relies heavily on past trajectories to prevent performance regression. If the underlying rules of an environment change over time, these historical trajectories may no longer represent correct behavior. Adapting this historical replay approach to rapidly changing environments remains a necessary direction for future research. Future evolution mechanisms could adopt time-weighted buffers to decay obsolete evidence.

\paragraph{Broader impacts.}
Skill-V supports more transparent agent adaptation by making skill revisions explicit and checking them against observed evidence. Its contract framework could be extended to incorporate safety requirements beyond task success. Future work could develop domain-specific constraints and shared evaluation protocols to assess whether these mechanisms help detect and mitigate harmful updates, with the broader goal of enabling trustworthy recursive self-improvement.

\section{Implementation Details}
\label{app:implementation}

This appendix describes the implementation of Skill-V in sufficient detail to separate the conceptual method from engineering choices. Skill-V maintains one training skill library and a synchronized evaluation copy. The training copy is updated exclusively using training trajectories. Before each validation round, the current training library is copied to the evaluation workers and remains strictly read-only throughout the validation. Rubric evaluations serve solely to diagnose and evolve the library. Unless explicitly stated otherwise, they are strictly excluded from the policy reward.

\subsection{End-to-End Training Procedure}
\label{app:training_algorithm}

The Algorithm~\ref{alg:skillv} summarizes the complete training loop. The same experience supports two complementary updates. Environment rewards optimize the policy, while terminal outcomes test the external skill library. Library evolution is performed every \(F\) training steps, with \(F=5\) in our experiments.

\begin{algorithm}[!t]
\caption{Skill-V training and library evolution}
\label{alg:skillv}
\begin{algorithmic}[1]
\Require Policy $\pi_{\theta}$, skill library $\mathcal{L}$, update interval $F$, retrieval budgets $K_g,K_t$
\For{training step $t=1,\ldots,T$}
    \Statex \textit{Skill-conditioned interaction}
    \State Sample task batch $\mathcal{X}_t$ and initialize $\mathcal{E}_t\gets\emptyset$
    \ForAll{$x\in\mathcal{X}_t$}
        \State $C(x)\gets\Call{SemanticRetrieve}{x,\mathcal{L},K_g,K_t}$
        \State $S(x)\gets\Call{ApplicabilityFilter}{x,C(x)}$
        \State $(\tau,y,r)\gets\Call{Rollout}{\pi_{\theta},x,S(x)}$
        \State $\mathcal{E}_t\gets\mathcal{E}_t\cup\{(x,\tau,y,r)\}$
    \EndFor
    \State Compute GRPO advantages from environment rewards $r$
    \If{$t\bmod F=0$}
        \Statex \textit{Outcome-grounded library evolution}
        \State $\mathcal{E}\gets$ recent training experiences
        \State $\mathcal{F}\gets\{e\in\mathcal{E}:y(e)=0\}$
        \State $\mathcal{P}_{\mathrm{new}}\gets\Call{ProposeNewSkills}{\mathcal{F},\mathcal{L}}$
        \State $\mathcal{L}\gets\Call{AddPendingSkills}{\mathcal{L},\mathcal{P}_{\mathrm{new}}}$
        \State $D\gets\Call{DiagnoseContracts}{\mathcal{L},\mathcal{E}}$
        \State $\widetilde{\mathcal{E}}\gets\Call{SelectEvidence}{D}$ \Comment{disagreements first}
        \State $\mathcal{P}\gets\Call{ProposeContractUpdates}{\mathcal{L},\widetilde{\mathcal{E}}}$
        \ForAll{$p\in\mathcal{P}$}
            \If{$\Call{ValidateContractUpdate}{p,\mathcal{L},\mathcal{E}}$}
                \State $\mathcal{L}\gets\Call{CommitVersion}{\mathcal{L},p}$
            \Else
                \State Retain the current contract version
            \EndIf
        \EndFor
    \EndIf
    \State $\theta\gets\Call{GRPOUpdate}{\theta,\mathcal{E}_t}$
\EndFor
\State \Return $\pi_{\theta}$ and $\mathcal{L}$
\end{algorithmic}
\end{algorithm}

Here, \textsc{DiagnoseContracts} executes every applicable contract possessing an available rubric and compares its judgment with the environment outcome. \textsc{SelectEvidence} prioritizes rubric-outcome disagreements while retaining supporting success and failure cases. \textsc{ProposeContractUpdates} serves two roles. It compiles rubrics for pending semantic skills and proposes targeted revisions to existing contracts. A newly added semantic skill becomes retrievable immediately. In contrast, a compiled or revised contract becomes active only after passing deterministic validation. \textsc{ValidateContractUpdate} combines schema and traceability checks, semantic preservation, non-empty executable support, and evidence non-regression. 

\subsection{Stored Skills as Versioned Contracts}
\label{app:contract_schema}

Free-form instructions alone do not distinguish what a skill means from how its execution can be checked. Skill-V therefore stores each skill as a versioned contract \(s_i=(\phi_i,p_i,\rho_i,\nu_i)\). Here, \(\phi_i\) specifies the semantic identity and intended scope, \(p_i\) provides the reusable procedure, \(\rho_i\) connects the skill to observable trajectory evidence, and \(\nu_i\) records its evolution. This explicit separation allows the executable boundary to be corrected without silently altering the core identity of the skill.

\begin{table*}[!t]
\centering
\small
\setlength{\tabcolsep}{6pt}
\renewcommand{\arraystretch}{1.18}
\caption{Fields of a stored Skill-V contract and their roles in skill use and evolution.}
\label{tab:contract-fields}
\begin{tabular}{p{0.25\textwidth}p{0.36\textwidth}p{0.35\textwidth}}
\toprule
\textbf{Component} & \textbf{Stored content} & \textbf{Role} \\
\midrule
\textbf{Semantic specification} \(\phi_i\)
& \texttt{skill\_id}, \texttt{title}, \texttt{principle}, \texttt{when\_to\_apply}, protected \texttt{skill\_constraints}
& Defines what capability the skill represents and establishes the semantic boundary that a revision is not allowed to overwrite. \\
\textbf{Reusable procedure} \(p_i\)
& Ordered natural-language steps
& Conditions the policy during interaction and may be refined when experience reveals an incomplete strategy. \\
\addlinespace[2pt]
\textbf{Executable rubric} \(\rho_i\)
& Applicability, required steps, forbidden behaviors, preconditions, postconditions, and termination conditions
& Translates the intended behavior into conservative criteria that can be checked deterministically against an executed trajectory. \\
\textbf{Evidence interface}
& Criterion identifier, severity, weight, match mode, source, operator, and pattern
& Grounds each executable criterion directly in the task, action, observation, or trajectory rather than in the policy's hidden reasoning. \\
\addlinespace[2pt]
\textbf{Evolution record} \(\nu_i\)
& Skill version, rubric version, traceability, update reason, and evidence case identifiers
& Links each committed version to its source skill and to the specific experience that motivated the change. \\
\bottomrule
\end{tabular}
\end{table*}

Table~\ref{tab:contract-examples} traces two revisions selected from the step-150 checkpoint of our main ALFWorld experiment. Each case follows the exact same evidence-to-update sequence. An environment outcome contradicts the current rubric, the executable boundary is modified while the skill intent remains fixed, and the candidate is replayed on the identical observed support before being committed.

\begin{table*}[!t]
\centering
\small
\setlength{\tabcolsep}{4pt}
\renewcommand{\arraystretch}{1.12}
\caption{How outcome evidence drives contract evolution in the main ALFWorld experiment at step 150. Skill-V tightens an under-specified boundary that admits failed behavior and relaxes an over-restrictive boundary that rejects successful behavior. Each candidate is evaluated against its previous version on identical replay support. The second revision is accepted as a non-regressive improvement, although it has not yet reached target thresholds.}
\label{tab:contract-examples}
\begin{tabular}{p{0.255\textwidth}c p{0.31\textwidth}c p{0.255\textwidth}}
\toprule
\rowcolor{skillvheader}
\multicolumn{5}{p{0.94\textwidth}}{\textbf{Case 1: Tightening an under-specified boundary}\hfill \texttt{gen\_004}: \emph{Track Counts \& Progress}} \\
\midrule
\rowcolor{skillvsemantic}
\multicolumn{5}{p{0.94\textwidth}}{\textbf{Skill intent, preserved:} Track how many goal objects remain and terminate only after the required count is reached.} \\
\cellcolor{skillvevolution}
\textbf{Counterexample}\newline
\emph{Contract: pass; outcome: failure.}\newline
Only one target object was collected in a quantity-two task.
&
\(\Longrightarrow\)
&
\cellcolor{skillvrubric}
\textbf{Rubric change}\newline
\textbf{Before:} no executable distinct-object count.\newline
\textbf{After:} require \(\geq 2\) \texttt{take} events on distinct objects with a hard \texttt{event\_count} criterion.
&
\(\Longrightarrow\)
&
\cellcolor{skillvsemantic}
\textbf{Replay validation}\newline
\begin{tabular}{@{}lrr@{}}
 & \textbf{Old} & \textbf{Revised}\\
Disc \(\downarrow\) & 0.400 & \textbf{0.100}\\
FNR \(\downarrow\)  & 1.000 & \textbf{0.000}\\
BAcc \(\uparrow\)   & 0.500 & \textbf{0.917}
\end{tabular}\newline
Support: \(10\rightarrow10\)\newline \textbf{Accepted}\newline
Skill v1\(\rightarrow\)v2; \newline  Rubric v1\(\rightarrow\)v2. \\
\midrule
\rowcolor{skillvheader}
\multicolumn{5}{p{0.94\textwidth}}{\textbf{Case 2: Relaxing an over-restrictive boundary}\hfill \texttt{gen\_008}: \emph{Avoid Redundant Rechecks}} \\
\midrule
\rowcolor{skillvsemantic}
\multicolumn{5}{p{0.94\textwidth}}{\textbf{Skill intent, preserved:} Record inspected locations and avoid revisiting them unless new evidence suggests a state change.} \\
\cellcolor{skillvevolution}
\textbf{Counterexample}\newline
\emph{Contract: fail; outcome: success.}\newline
A successful trajectory contained a brief pair of identical consecutive actions.
&
\(\Longrightarrow\)
&
\cellcolor{skillvrubric}
\textbf{Rubric change}\newline
\textbf{Before:} flag two identical consecutive actions as a loop.\newline
\textbf{After:} preserve loop detection but trigger it after three identical consecutive actions.
&
\(\Longrightarrow\)
&
\cellcolor{skillvsemantic}
\textbf{Replay validation}\newline
\begin{tabular}{@{}lrr@{}}
 & \textbf{Old} & \textbf{Revised}\\
Disc \(\downarrow\) & 0.500 & \textbf{0.333}\\
FNR \(\downarrow\)  & 1.000 & \textbf{0.667}\\
BAcc \(\uparrow\)   & 0.500 & \textbf{0.667}
\end{tabular}\newline
Support: \(12\rightarrow12\)\ \newline  \textbf{Accepted}\newline
Skill v1\(\rightarrow\)v2;\newline Rubric v2\(\rightarrow\)v3. \\
\bottomrule
\end{tabular}
\end{table*}

The first revision compiles a semantic multiplicity requirement into an executable distinct-object counter. The second revision utilizes successful replay as a counterexample against an over-restrictive loop detector. These cases demonstrate that rubric evolution is not equivalent to simply accumulating constraints. Instead, it dynamically calibrates the executable boundary in either direction and commits a new version only when replay evidence proves strictly non-regressive.

The checked-in initial libraries utilize a compact sentence-rubric schema for readability. At load time, this representation is normalized into the internal criterion groups outlined in Table~\ref{tab:contract-fields}. Each executable criterion contains a stable identifier, a description, a severity in \(\{\textsc{Hard},\textsc{Soft}\}\), a weight in \([0,1]\), a match mode in \(\{\textsc{Any},\textsc{All}\}\), and at least one observable evidence matcher. Hard criteria employ unit weight. Protected semantic constraints reside within the same contract but are never treated as executable evidence unless a conservative proxy has been explicitly compiled for them.

To ensure strict security and stability, Skill-V never executes model-generated code. Evidence matchers remain restricted to a predefined declarative operator set. All domains support case-insensitive literal matching and regular expressions over the task, executed actions, environment observations, or their flattened trajectory. ALFWorld additionally supports bounded event-sequence and event-count monitors. The flattened trajectory contains only actions and observations. Policy thoughts are strictly excluded, ensuring that a planned but unexecuted action cannot fraudulently satisfy a rubric criterion.

\subsection{Executable Rubric Evaluation}
\label{app:rubric_execution}

For a given task \(x\), the verifier first evaluates the declarative applicability conditions of each contract. Empty applicability conditions are treated as universally applicable. Invalid generated contracts fail closed and are automatically excluded from the policy prompt. If a valid contract is deemed applicable, the verifier systematically evaluates its execution criteria against trajectory \(\tau\).

Let \(m_{ij}(\tau)\in\{0,1\}\) indicate whether criterion \(c_{ij}\) is satisfied after accounting for its criterion type. In particular, a forbidden-behavior criterion passes when its specified evidence pattern remains absent. Let \(w_{ij}\) denote its weight. Across all applicable contracts possessing executable criteria, Skill-V computes
\begin{equation}
q(\tau)=
\frac{\sum_{s_i\in\mathcal{A}(x)}\sum_j w_{ij}m_{ij}(\tau)}
{\sum_{s_i\in\mathcal{A}(x)}\sum_j w_{ij}},
\qquad
\hat y^{\mathrm{rub}}(\tau)=
\mathbb{I}\!\left[q(\tau)\geq\eta\ \land\ H(\tau)=0\right],
\label{eq:app-rubric-aggregate}
\end{equation}
where \(H(\tau)\) indicates whether any applicable hard criterion is violated. We utilize a threshold of \(\eta=0.7\). When no applicable contract possesses an executable criterion, the aggregate score and judgment are undefined, and the trajectory generates no rubric-outcome evidence.

In addition to the aggregate judgment, the implementation logs per-skill applicability, pass/fail state, hard-criterion state, weighted score, and failed criterion identifiers. These diagnostics localize candidate revisions. They are deliberately not interpreted as causal credit assignment, because the environment outcome is evaluated for the complete trajectory rather than attributed to an individual skill.

\subsection{Failure-Driven Skill Acquisition}
\label{app:skill_acquisition}

At each library-update step, the acquisition updater receives failed training trajectories alongside the titles of existing skills. It considers at most five failure examples and retains the final five interaction steps of each example to provide context. The external language model then proposes up to a configured maximum number of reusable skills. Each proposal contains a title, principle, and intended scope. Existing titles serve as a deduplication hint, and the implementation assigns collision-free identifiers locally.

A newly acquired skill is immediately converted into a semantic contract. Its principle serves as its initial procedure, and an immutable semantic constraint is instantiated with the status \(\textsc{PendingCompilation}\). At this point, the skill can participate in embedding retrieval, applicability judgment, and policy prompting. However, it cannot produce a rubric judgment because it lacks executable criteria. This strict separation prevents an uncalibrated model-generated rule from acting as a verifier.

Pending skills receive prioritization by the contract updater. A successful compilation adds conservative executable proxies and transitions the skill into its first executable contract. A failed or rejected compilation retains the pending status, allowing the semantic skill to remain available to the policy for subsequent interactions.

\subsection{Outcome-Grounded Contract Diagnosis}
\label{app:contract_diagnosis}

For each recent training experience \((x,\tau,y)\), Skill-V evaluates all valid contracts whose declarative applicability conditions match \(x\). Comparing the aggregate rubric judgment with the environment outcome produces the four diagnostic cases detailed in Table~\ref{tab:evidence-quadrants}.

\begin{table}[htbp]
\centering
\small
\caption{Interpretation of rubric-outcome evidence.}
\label{tab:evidence-quadrants}
\begin{tabular}{ccc p{0.48\linewidth}}
\toprule
Rubric & Outcome & Agreement & Interpretation \\
\midrule
Pass & Success & Yes & Evidence supporting the current operational boundary. \\
Fail & Failure & Yes & Consistent negative evidence. \\
Pass & Failure & No  & The contract may be too permissive or under-specified. \\
Fail & Success & No  & The contract may be too restrictive. \\
\bottomrule
\end{tabular}
\end{table}

The updater selects at most 12 recent cases, prioritizing the two disagreement quadrants while retaining agreement cases from both outcome classes. It then retrieves up to 24 relevant contracts to form the proposal context. Relevance favors contracts implicated by failed criteria, contracts whose semantic fields overlap the current task, and pending contracts awaiting compilation.

The external updater may refine applicability conditions, procedural steps, or executable criteria. For an existing skill, however, every protected semantic constraint must be copied perfectly. Consequently, a revision can successfully repair the operational boundary of a skill, but it remains structurally prohibited from silently replacing its semantic identity.

\subsection{Deterministic Revision Validation}
\label{app:revision_validation}

Language-model proposals are never committed directly. Skill-V first verifies the contract schema, unique criterion identifiers, allowed matcher types, criterion weights and severities, criterion-to-procedure traceability, source skill identity, and monotonically increasing versions. Following structural validation, it performs shadow replay on the observed evidence pool associated with the update.

For a skill \(s_i\), rubric pass/fail is evaluated as a prediction of the environment outcome. With the standard definitions of true positives (TP), false positives (FP), false negatives (FN), and true negatives (TN), we compute
\begin{equation}
\operatorname{Disc}=\frac{\mathrm{FP}+\mathrm{FN}}{N},\qquad
\operatorname{FNR}=\frac{\mathrm{FN}}{\mathrm{TP}+\mathrm{FN}},\qquad
\operatorname{BAcc}=\frac{1}{2}\left(
\frac{\mathrm{TP}}{\mathrm{TP}+\mathrm{FN}}+
\frac{\mathrm{TN}}{\mathrm{TN}+\mathrm{FP}}
\right).
\label{eq:app-revision-metrics}
\end{equation}

Let \(\rho_i\) and \(\rho_i'\) denote the current and proposed rubrics. A proposal is considered evidence-preserving only if
\begin{equation}
\operatorname{Disc}(\rho_i')\leq\operatorname{Disc}(\rho_i),\quad
\operatorname{FNR}(\rho_i')\leq\operatorname{FNR}(\rho_i),\quad
\operatorname{BAcc}(\rho_i')\geq\operatorname{BAcc}(\rho_i).
\label{eq:app-revision-gate}
\end{equation}
It must additionally possess non-empty executable support. If a hard executable criterion changes, at least one quantity in Eq.~\ref{eq:app-revision-gate} must improve strictly. The target values used to prompt the updater (\(\operatorname{FNR}\leq0.05\) and \(\operatorname{Disc}\leq0.15\)) merely indicate whether an accepted proposal reaches a desired operating region. They do not constitute additional hard acceptance requirements.

The implementation robustly handles degenerate evidence pools using standard conventions. When no valid applicable and executable trajectory exists, support is zero, disagreement and FNR default to one, and balanced accuracy defaults to zero. The candidate is subsequently rejected by the non-empty-support condition. If the evidence contains no positive outcomes, FNR is zero and the missing true-positive rate is treated as one. If it contains no negative outcomes, the missing true-negative rate is treated as one. Current and candidate contracts are replayed over the exact same observed trajectory pool, although their effective supports may diverge when a revision alters applicability.

An accepted proposal increments both the skill and rubric versions and replaces the current contract in place. A rejected proposal leaves the current version completely unchanged. Because revision occurs in place, accepted revisions fundamentally do not increase library size.

\subsection{Semantic Retrieval and Applicability Filtering}
\label{app:retrieval_filtering}

Skill-V embeds the task instruction and each skill's semantic representation using Qwen3-Embedding-0.6B. General and task-specific skills are ranked separately. For final budgets \(K_g=6\) and \(K_t=5\), the implementation first forms candidate pools of size \(\lceil\alpha K_g\rceil\) and \(\lceil\alpha K_t\rceil\), where \(\alpha=2\).

An external language model labels each candidate as \(\textsc{Applicable}\), \(\textsc{NotApplicable}\), or \(\textsc{Uncertain}\), returning this label alongside a confidence score. Only \(\textsc{NotApplicable}\) predictions with confidence at least \(\delta=0.8\) are removed. Applicable and uncertain candidates remain fully eligible. Among these remaining candidates, Skill-V removes those whose cosine similarity falls more than \(\epsilon=0.15\) below the highest remaining score, ultimately selecting up to the original category-specific budget.

The judge must return one valid decision for every supplied candidate. If an API request fails or returns an incomplete decision set after two attempts, the system gracefully falls back to the original embedding Top-\(K\) result. Decisions are cached using the normalized task, model version, prompt version, skill identifier, skill and rubric versions, and a fingerprint of the semantic fields. Updating a skill therefore invalidates stale applicability decisions naturally through the cache key.

\section{Experimental Details}
\label{app:experimental_details}

\subsection{Environments and Metrics}
\label{app:environments}

We evaluate Skill-V in two interactive domains. ALFWorld contains embodied household tasks and reports success rates for Pick, Look, Clean, Heat, Cool, Pick2, and the complete task set. Its binary outcome is derived from a positive terminal task reward. WebShop requires an agent to search for and purchase a product satisfying a natural-language request. We report its average task score on a 0--100 scale alongside its overall success rate. For contract diagnosis, a WebShop trajectory is labeled successful only when its normalized environment score reaches at least \(0.99\).

All headline results in \Cref{tab:main_results} correspond to the selected step-150 checkpoints. We deliberately avoid choosing a separate peak checkpoint for each reported metric. The main table consequently reports a \(95.3\%\) overall success rate for ALFWorld, and an average score of \(92.5\) with an \(85.9\%\) success rate for WebShop.

\subsection{Training and Evaluation Configuration}
\label{app:configuration}

Both domains employ GRPO with eight rollouts per instruction, a learning rate of \(10^{-6}\), a KL loss coefficient of \(0.01\), and an invalid-action penalty of \(0.1\). The policy is initialized from the corresponding 7B SFT checkpoint and trained across eight GPUs. Table~\ref{tab:domain-config} lists domain-specific settings. These settings reflect the effective launch configuration of the reported runs, incorporating any subsequent command-line overrides.

\begin{table*}[htbp]
\centering
\small
\caption{Domain-specific configuration for the reported step-150 results. Update limits refer to per-library-update event budgets rather than global limits.}
\label{tab:domain-config}
\begin{tabular}{lcc}
\toprule
Configuration & ALFWorld & WebShop \\
\midrule
Training batch size & 16 & 32 \\
Validation batch size & 64 & 128 \\
Rollouts per instruction & 8 & 8 \\
Maximum environment steps & 50 & 15 \\
Maximum prompt length & 8192 & 6000 \\
Maximum response length & 512 & 768 \\
Validation sampling & sampling & greedy \\
Validation temperature & 0.4 & 0.4 (sampling disabled) \\
Library update frequency & 5 steps & 5 steps \\
Maximum new semantic skills & 10 & 3 \\
Maximum contract revisions & 10 & 3 \\
Rubric pass threshold \(\eta\) & 0.7 & 0.7 \\
Outcome success threshold & positive task reward & normalized score \(\geq0.99\) \\
General/task-specific Top-\(K\) & 6/5 & 6/5 \\
Candidate multiplier \(\alpha\) & 2.0 & 2.0 \\
Applicability confidence \(\delta\) & 0.8 & 0.8 \\
Embedding margin \(\epsilon\) & 0.15 & 0.15 \\
Tensor parallel size & 4 & 4 \\
Checkpoint/validation interval & 10/5 steps & 10/5 steps \\
Rubric process reward & disabled & disabled \\
\bottomrule
\end{tabular}
\end{table*}

\subsection{Initial Skill Libraries}
\label{app:initial_libraries}

The initial ALFWorld library contains 12 general skills, 32 task-specific skills, and 11 common mistakes. The 44 general and task-specific entries function as retrievable contracts. Common mistakes are stored separately, with up to five randomly inserted into the prompt as negative guidance. The initial WebShop library contains 15 general skills, 39 task-specific skills, and 12 common mistakes. Thus, the complete initial stored-library sizes total 55 and 66 entries, respectively.

The initial contracts were compiled offline from previously generated procedural skills. Unsupported semantic requirements remain protected constraints but are disabled as executable proxies. This prevents the initial library from erroneously presenting a deterministic matcher as complete verification of a richer semantic claim.

\subsection{External Models}
\label{app:external_models}

Qwen3-Embedding-0.6B is utilized exclusively for semantic retrieval. The reported runs employ \texttt{deepseek-v4-flash} as the external language model for applicability judgment, failure-driven skill proposals, and rubric compilation or revision. The policy itself acts as a separate 7B actor and remains the only model updated by gradient-based reinforcement learning. External-model responses are parsed as JSON and verified locally. The updater structurally cannot bypass the contract validator or the evidence-preservation gate.

The applicability judge operates under a limit of 2,048 completion tokens, two attempts, and a 30-second request timeout. A five-minute circuit-breaker cooldown avoids rapid repeated requests after service failures. Semantic acquisition and contract maintenance utilize limits of 4,096 and 8,192 completion tokens, respectively, with two parsing attempts.


\section{Library Evolution and Diagnostics}
\label{app:library_analysis}

\subsection{Library Evolution Statistics}
\label{app:library_statistics}

\begin{table}[!t]
\centering
\small
\caption{Skill-library evolution through step 150. Total entries include common mistakes, which are stored separately from retrievable contracts.}
\label{tab:library-stats}
\begin{tabular}{lrrrr}
\toprule
Domain & Initial & Step 150 & Net Growth & Accepted Revisions \\
\midrule
ALFWorld & 55 & 205 & 150 & 10 \\
WebShop & 66 & 153 & 87 & 14 \\
\bottomrule
\end{tabular}
\end{table}
Table~\ref{tab:library-stats} reports the library state at the selected checkpoints. Because revisions replace existing versions in place, they fundamentally do not contribute to size growth.
For ALFWorld, the step-150 library contains 162 general skills, 32 task-specific skills, and 11 common mistakes. The 150-entry net growth arises entirely from newly acquired semantic skills. Ten contract revisions were successfully accepted by step 150 at update steps 5, 60, 65, 85, 120, 125, and 150. The corresponding numbers of accepted revisions were 2, 1, 2, 1, 1, 1, and 2. This exact count is confirmed independently by both the summed training metric and the version history stored within the checkpoint.

WebShop contains 153 stored entries at step 150 and records 14 accepted revisions. Its net growth is computed purely from the final and initial library sizes, as the revisions themselves execute as in-place updates. These robust statistics demonstrate that the reported method does far more than simply grow its library. It actively calibrates the operational boundaries of persistent skills as empirical evidence accumulates.

\subsection{Checkpoint Diagnostics}
\label{app:checkpoint_diagnostics}

The ALFWorld training log retains additional aggregate diagnostics for the selected checkpoint (Table~\ref{tab:alfworld-diagnostics}). These metrics characterize the final training state but do not substitute for the standard task-success metrics utilized for comparison with prior work.

\begin{table}[htbp]
\centering
\small
\caption{Aggregate ALFWorld diagnostics at the selected step-150 checkpoint.}
\label{tab:alfworld-diagnostics}
\begin{tabular}{lr}
\toprule
Metric & Value \\
\midrule
Validation success (\%) & 95.3 \\
Validation task score & 8.130 \\
Training-episode success (\%) & 86.7 \\
Validation rubric score & 0.894 \\
Rubric-outcome disagreement & 0.033 \\
Hard-criterion violation rate & 0.062 \\
Stored library entries & 205 \\
\bottomrule
\end{tabular}
\end{table}

The extremely low aggregate disagreement value indicates that executable rubric judgments remain highly consistent with environment outcomes at this checkpoint. This value serves a descriptive purpose rather than acting as a causal measure of policy quality, as the outcome is trajectory-level and several contracts may apply simultaneously to the same task.

\section{Prompt Templates}
\label{app:prompts}

This section records the effective instructions supplied to the external language models. Runtime prompts dynamically substitute the task, candidate skills, trajectory cases, diagnostics, and current contracts into the bracketed fields.

\subsection{Applicability Judgment}

\begin{quote}\small
Decide whether each candidate skill is semantically applicable to the task objective. Applicability means the skill could be useful for solving this task at some point in the episode. Judge only the task objective and declared skill scope. Do not require that a precondition already holds in the current environment state. Handle paraphrases and implied state changes.

Use \texttt{applicable} when the skill is clearly compatible, \texttt{not\_applicable} when it is clearly incompatible, and \texttt{uncertain} when compatibility cannot be established confidently. Return every supplied skill identifier exactly once as a JSON object containing the decision, confidence in \([0,1]\), and a short analytical reason.

\textbf{[Input Fields]} Task objective; Candidate skills (ID, Title, Principle, Scope).
\end{quote}

\subsection{Failure-Driven Semantic Acquisition}

\begin{quote}\small
Analyze the supplied failed trajectories and suggest new skills to add to the skill library. Avoid duplicating existing skill titles. Generate between one and the configured maximum number of actionable, reusable skills that could prevent these failures. Each skill must contain a title, a brief principle, and an explicit condition describing when it should be applied. Return only a JSON object containing a \texttt{skills} array.

\textbf{[Input Fields]} Failed trajectories (up to five failures, final five steps each); Existing skill titles.
\end{quote}

\subsection{Contract Compilation and Revision}

\begin{quote}\small
Maintain a skill library under explicit hybrid contracts. First derive semantic constraints from the skill principle and make every constraint traceable to the procedure. Then add conservative executable proxies using safe, domain-appropriate matchers. Never claim that a proxy completely verifies a richer semantic constraint.

Use rubric-fail/outcome-fail cases to repair a skill while preserving valid criteria. A rubric-pass/outcome-fail case indicates a missed failure and may require an added or conditionalized criterion. A rubric-fail/outcome-pass case indicates an over-restrictive rubric and may require narrower applicability or relaxed soft criteria. Preserve a supported contract unless repeated evidence justifies a change. For any revision, copy every protected semantic constraint exactly. Prefer revising a relevant contract over adding a duplicate. Return only JSON that perfectly follows the supplied contract schema and update limits.

\textbf{[Input Fields]} Trajectory cases; Per-skill rubric diagnostics; Relevant existing contracts.
\end{quote}


%% file: main.bbl
\begin{thebibliography}{35}
\providecommand{\natexlab}[1]{#1}
\providecommand{\url}[1]{\texttt{#1}}
\expandafter\ifx\csname urlstyle\endcsname\relax
  \providecommand{\doi}[1]{doi: #1}\else
  \providecommand{\doi}{doi: \begingroup \urlstyle{rm}\Url}\fi

\bibitem[Anthropic()]{anthropic}
Anthropic.
\newblock The claude 3 model family: Opus, sonnet, haiku.
\newblock URL \url{https://api.semanticscholar.org/CorpusID:270640496}.

\bibitem[Chen et~al.(2026)Chen, Shi, Yang, and Xu]{SkillReranker}
Yanping Chen, Weijie Shi, Wen Yang, and Jiajie Xu.
\newblock Task decomposition-guided reranking for adaptive agent skill retrieval.
\newblock \emph{CoRR}, abs/2607.06283, 2026.
\newblock \doi{10.48550/ARXIV.2607.06283}.
\newblock URL \url{https://doi.org/10.48550/arXiv.2607.06283}.

\bibitem[Chhikara et~al.(2025)Chhikara, Khant, Aryan, Singh, and Yadav]{Mem0}
Prateek Chhikara, Dev Khant, Saket Aryan, Taranjeet Singh, and Deshraj Yadav.
\newblock Mem0: Building production-ready {AI} agents with scalable long-term memory.
\newblock In In{\^{e}}s Lynce, Nello Murano, Mauro Vallati, Serena Villata, Federico Chesani, Michela Milano, Andrea Omicini, and Mehdi Dastani (eds.), \emph{{ECAI} 2025 - 28th European Conference on Artificial Intelligence, 25-30 October 2025, Bologna, Italy - Including 14th Conference on Prestigious Applications of Intelligent Systems {(PAIS} 2025)}, volume 413 of \emph{Frontiers in Artificial Intelligence and Applications}, pp.\  2993--3000. {IOS} Press, 2025.
\newblock \doi{10.3233/FAIA251160}.
\newblock URL \url{https://doi.org/10.3233/FAIA251160}.

\bibitem[DeepSeek{-}AI(2025)]{deepseek}
DeepSeek{-}AI.
\newblock Deepseek-v3.2: Pushing the frontier of open large language models.
\newblock \emph{CoRR}, abs/2512.02556, 2025.
\newblock \doi{10.48550/ARXIV.2512.02556}.
\newblock URL \url{https://doi.org/10.48550/arXiv.2512.02556}.

\bibitem[French(1999)]{french1999catastrophic}
Robert~M French.
\newblock Catastrophic forgetting in connectionist networks.
\newblock \emph{Trends in cognitive sciences}, 3\penalty0 (4):\penalty0 128--135, 1999.

\bibitem[Gou et~al.(2024)Gou, Shao, Gong, Shen, Yang, Duan, and Chen]{CRITIC_iclr}
Zhibin Gou, Zhihong Shao, Yeyun Gong, Yelong Shen, Yujiu Yang, Nan Duan, and Weizhu Chen.
\newblock {CRITIC:} large language models can self-correct with tool-interactive critiquing.
\newblock In \emph{The Twelfth International Conference on Learning Representations, {ICLR} 2024, Vienna, Austria, May 7-11, 2024}. OpenReview.net, 2024.
\newblock URL \url{https://openreview.net/forum?id=Sx038qxjek}.

\bibitem[Liu et~al.(2026{\natexlab{a}})Liu, Su, Xia, Han, Zheng, Xie, Ding, and Yao]{SimpleMem}
Jiaqi Liu, Yaofeng Su, Peng Xia, Siwei Han, Zeyu Zheng, Cihang Xie, Mingyu Ding, and Huaxiu Yao.
\newblock Simplemem: Efficient lifelong memory for {LLM} agents.
\newblock \emph{CoRR}, abs/2601.02553, 2026{\natexlab{a}}.
\newblock \doi{10.48550/ARXIV.2601.02553}.
\newblock URL \url{https://doi.org/10.48550/arXiv.2601.02553}.

\bibitem[Liu et~al.(2026{\natexlab{b}})Liu, Lin, Zhang, Ren, Wu, Li, Wang, Fu, and Ye]{liu2026pathrsi}
Shuaiqi Liu, Zhengkai Lin, Yuxiang Zhang, Yuanyi Ren, Yue Wu, Yongbin Li, Zheng Wang, Zhihang Fu, and Jieping Ye.
\newblock The path to recursive self-improving agents: Foundation, framework, and future directions.
\newblock 2026{\natexlab{b}}.

\bibitem[Liu et~al.(2026{\natexlab{c}})Liu, Su, Xie, Zhang, Zong, Guo, Xie, Ji, Yim, Luo, Ren, Chenyu, Li, and Song]{skillrevise}
Yuxuan Liu, Zhaochen Su, Lingyun Xie, Yuhao Zhang, Qing Zong, Jiahe Guo, Zhongwei Xie, Yiyan Ji, Yauwai Yim, Hongyu Luo, Xiyu Ren, Ruan Chenyu, Haoran Li, and Yangqiu Song.
\newblock Skillrevise: Improving llm-authored agent skills via trace-conditioned skill revision.
\newblock \emph{CoRR}, abs/2606.01139, 2026{\natexlab{c}}.
\newblock \doi{10.48550/ARXIV.2606.01139}.
\newblock URL \url{https://doi.org/10.48550/arXiv.2606.01139}.

\bibitem[OpenAI(2023)]{openai}
OpenAI.
\newblock {GPT-4} technical report.
\newblock \emph{CoRR}, abs/2303.08774, 2023.
\newblock \doi{10.48550/ARXIV.2303.08774}.
\newblock URL \url{https://doi.org/10.48550/arXiv.2303.08774}.

\bibitem[Ouyang et~al.(2026)Ouyang, Yan, Chen, Han, Wang, Mishra, Meng, Li, Jiao, Zha, Shen, Tirumalashetty, Lee, Han, Pfister, and Lee]{skillos}
Siru Ouyang, Jun Yan, Yanfei Chen, Rujun Han, Zifeng Wang, Bhavana~Dalvi Mishra, Rui Meng, Chun{-}Liang Li, Yizhu Jiao, Kaiwen Zha, Maohao Shen, Vishy Tirumalashetty, George Lee, Jiawei Han, Tomas Pfister, and Chen{-}Yu Lee.
\newblock Skillos: Learning skill curation for self-evolving agents.
\newblock \emph{CoRR}, abs/2605.06614, 2026.
\newblock \doi{10.48550/ARXIV.2605.06614}.
\newblock URL \url{https://doi.org/10.48550/arXiv.2605.06614}.

\bibitem[Packer et~al.(2023)Packer, Wooders, Lin, Fang, Patil, Stoica, and Gonzalez]{MemGPT}
Charles Packer, Sarah Wooders, Kevin Lin, Vivian Fang, Shishir~G. Patil, Ion Stoica, and Joseph Gonzalez.
\newblock Memgpt: Towards llms as operating systems.
\newblock \emph{ArXiv}, abs/2310.08560, 2023.
\newblock URL \url{https://api.semanticscholar.org/CorpusID:263909014}.

\bibitem[Pu et~al.(2026)Pu, Song, and Zhao]{skillops}
Hongji Pu, Xinyuan Song, and Liang Zhao.
\newblock Skillops: Managing {LLM} agent skill libraries as self-maintaining software ecosystems.
\newblock \emph{CoRR}, abs/2605.13716, 2026.
\newblock \doi{10.48550/ARXIV.2605.13716}.
\newblock URL \url{https://doi.org/10.48550/arXiv.2605.13716}.

\bibitem[Schick et~al.(2023)Schick, Dwivedi-Yu, Dessi, Raileanu, Lomeli, Hambro, Zettlemoyer, Cancedda, and Scialom]{Toolformer}
Timo Schick, Jane Dwivedi-Yu, Roberto Dessi, Roberta Raileanu, Maria Lomeli, Eric Hambro, Luke Zettlemoyer, Nicola Cancedda, and Thomas Scialom.
\newblock Toolformer: Language models can teach themselves to use tools.
\newblock In A.~Oh, T.~Naumann, A.~Globerson, K.~Saenko, M.~Hardt, and S.~Levine (eds.), \emph{Advances in Neural Information Processing Systems}, volume~36, pp.\  68539--68551. Curran Associates, Inc., 2023.
\newblock \doi{10.52202/075280-2997}.
\newblock URL \url{https://proceedings.neurips.cc/paper_files/paper/2023/file/d842425e4bf79ba039352da0f658a906-Paper-Conference.pdf}.

\bibitem[Shi et~al.(2026)Shi, Chen, Lu, Miao, Liu, Gu, Cai, Wang, and Zhang]{skill1}
Yaorui Shi, Yuxin Chen, Zhengxi Lu, Yuchun Miao, Shugui Liu, Qi~Gu, Xunliang Cai, Xiang Wang, and An~Zhang.
\newblock Skill1: Unified evolution of skill-augmented agents via reinforcement learning.
\newblock \emph{CoRR}, abs/2605.06130, 2026.
\newblock \doi{10.48550/ARXIV.2605.06130}.
\newblock URL \url{https://doi.org/10.48550/arXiv.2605.06130}.

\bibitem[Shinn et~al.(2023)Shinn, Cassano, Gopinath, Narasimhan, and Yao]{Reflexion}
Noah Shinn, Federico Cassano, Ashwin Gopinath, Karthik Narasimhan, and Shunyu Yao.
\newblock Reflexion: language agents with verbal reinforcement learning.
\newblock In Alice Oh, Tristan Naumann, Amir Globerson, Kate Saenko, Moritz Hardt, and Sergey Levine (eds.), \emph{Advances in Neural Information Processing Systems 36: Annual Conference on Neural Information Processing Systems 2023, NeurIPS 2023, New Orleans, LA, USA, December 10 - 16, 2023}, 2023.
\newblock URL \url{http://papers.nips.cc/paper\_files/paper/2023/hash/1b44b878bb782e6954cd888628510e90-Abstract-Conference.html}.

\bibitem[Shridhar et~al.(2020)Shridhar, Yuan, C{\^{o}}t{\'{e}}, Bisk, Trischler, and Hausknecht]{ALFWorld}
Mohit Shridhar, Xingdi Yuan, Marc{-}Alexandre C{\^{o}}t{\'{e}}, Yonatan Bisk, Adam Trischler, and Matthew~J. Hausknecht.
\newblock Alfworld: Aligning text and embodied environments for interactive learning.
\newblock \emph{CoRR}, abs/2010.03768, 2020.
\newblock URL \url{https://arxiv.org/abs/2010.03768}.

\bibitem[Team(2023)]{gemini}
Gemini Team.
\newblock Gemini: {A} family of highly capable multimodal models.
\newblock \emph{CoRR}, abs/2312.11805, 2023.
\newblock \doi{10.48550/ARXIV.2312.11805}.
\newblock URL \url{https://doi.org/10.48550/arXiv.2312.11805}.

\bibitem[Team(2025)]{kimi}
Kimi Team.
\newblock Kimi {K2:} open agentic intelligence.
\newblock \emph{CoRR}, abs/2507.20534, 2025.
\newblock \doi{10.48550/ARXIV.2507.20534}.
\newblock URL \url{https://doi.org/10.48550/arXiv.2507.20534}.

\bibitem[Vishe et~al.(2026)Vishe, Surana, Jiang, Huang, Li, Kuang, Yu, Rossi, Shang, McAuley, and Wu]{Skill-R1}
Yash Vishe, Rohan Surana, Xunyi Jiang, Zihan Huang, Xintong Li, Nikki~Lijing Kuang, Tong Yu, Ryan~A. Rossi, Jingbo Shang, Julian~J. McAuley, and Junda Wu.
\newblock Skill-r1: Agent skill evolution via reinforcement learning.
\newblock \emph{CoRR}, abs/2605.09359, 2026.
\newblock \doi{10.48550/ARXIV.2605.09359}.
\newblock URL \url{https://doi.org/10.48550/arXiv.2605.09359}.

\bibitem[Wang et~al.(2024{\natexlab{a}})Wang, Xie, Jiang, Mandlekar, Xiao, Zhu, Fan, and Anandkumar]{Voyager}
Guanzhi Wang, Yuqi Xie, Yunfan Jiang, Ajay Mandlekar, Chaowei Xiao, Yuke Zhu, Linxi Fan, and Anima Anandkumar.
\newblock Voyager: An open-ended embodied agent with large language models.
\newblock \emph{Trans. Mach. Learn. Res.}, 2024, 2024{\natexlab{a}}.
\newblock URL \url{https://openreview.net/forum?id=ehfRiF0R3a}.

\bibitem[Wang et~al.(2026)Wang, Yan, Wang, Tian, Mishra, Xu, Gandhi, Xu, and Cheong]{SAGE_Skill}
Jiongxiao Wang, Qiaojing Yan, Yawei Wang, Yijun Tian, Soumya~Smruti Mishra, Zhichao Xu, Megha Gandhi, Panpan Xu, and Lin~Lee Cheong.
\newblock Reinforcement learning for self-improving agent with skill library.
\newblock In Maria Liakata, Viviane~P. Moreira, Jiajun Zhang, and David Jurgens (eds.), \emph{Proceedings of the 64th Annual Meeting of the Association for Computational Linguistics (Volume 1: Long Papers), {ACL} 2026, San Diego, California, United States, July 2-7, 2026}, pp.\  1529--1550. Association for Computational Linguistics, 2026.
\newblock \doi{10.18653/V1/2026.ACL-LONG.69}.
\newblock URL \url{https://doi.org/10.18653/v1/2026.acl-long.69}.

\bibitem[Wang et~al.(2024{\natexlab{b}})Wang, Zhang, Su, and Zhu]{wang2023continual}
Liyuan Wang, Xingxing Zhang, Hang Su, and Jun Zhu.
\newblock A comprehensive survey of continual learning: Theory, method and application.
\newblock \emph{IEEE Transactions on Pattern Analysis and Machine Intelligence}, 46\penalty0 (8):\penalty0 5362--5383, 2024{\natexlab{b}}.
\newblock \doi{10.1109/TPAMI.2024.3367329}.

\bibitem[Wu et~al.(2025)Wu, Wang, Mei, Cai, Fu, Yang, Wen, Yang, Shen, Wang, and Shi]{EvolveR}
Rong Wu, Xiaoman Wang, Jianbiao Mei, Pinlong Cai, Daocheng Fu, Cheng Yang, Licheng Wen, Xuemeng Yang, Yufan Shen, Yuxin Wang, and Botian Shi.
\newblock Evolver: Self-evolving {LLM} agents through an experience-driven lifecycle.
\newblock \emph{CoRR}, abs/2510.16079, 2025.
\newblock \doi{10.48550/ARXIV.2510.16079}.
\newblock URL \url{https://doi.org/10.48550/arXiv.2510.16079}.

\bibitem[Xia et~al.(2026)Xia, Chen, Wang, Liu, Zeng, Wang, Han, Zhou, Zhao, Chen, Zheng, Xie, and Yao]{skillrl}
Peng Xia, Jianwen Chen, Hanyang Wang, Jiaqi Liu, Kaide Zeng, Yu~Wang, Siwei Han, Yiyang Zhou, Xujiang Zhao, Haifeng Chen, Zeyu Zheng, Cihang Xie, and Huaxiu Yao.
\newblock Skillrl: Evolving agents via recursive skill-augmented reinforcement learning.
\newblock \emph{CoRR}, abs/2602.08234, 2026.
\newblock \doi{10.48550/ARXIV.2602.08234}.
\newblock URL \url{https://doi.org/10.48550/arXiv.2602.08234}.

\bibitem[Yang et~al.(2024)Yang, Wang, Lu, Liu, Le, Zhou, and Chen]{opro}
Chengrun Yang, Xuezhi Wang, Yifeng Lu, Hanxiao Liu, Quoc~V Le, Denny Zhou, and Xinyun Chen.
\newblock Large language models as optimizers.
\newblock 2024.

\bibitem[Yang et~al.(2026)Yang, Gong, Huang, Yang, Zhou, Huang, Li, Gao, Dai, Liu, Qiu, Yang, Chen, Yang, and Luo]{skillopt2026}
Yifan Yang, Ziyang Gong, Weiquan Huang, Qihao Yang, Ziwei Zhou, Zisu Huang, Yan Li, Xuemei Gao, Qi~Dai, Bei Liu, Kai Qiu, Yuqing Yang, Dongdong Chen, Xue Yang, and Chong Luo.
\newblock Skillopt: Executive strategy for self-evolving agent skills, 2026.
\newblock URL \url{https://arxiv.org/abs/2605.23904}.

\bibitem[Yao et~al.(2022)Yao, Chen, Yang, and Narasimhan]{webshop}
Shunyu Yao, Howard Chen, John Yang, and Karthik Narasimhan.
\newblock Webshop: Towards scalable real-world web interaction with grounded language agents.
\newblock In Sanmi Koyejo, S.~Mohamed, A.~Agarwal, Danielle Belgrave, K.~Cho, and A.~Oh (eds.), \emph{Advances in Neural Information Processing Systems 35: Annual Conference on Neural Information Processing Systems 2022, NeurIPS 2022, New Orleans, LA, USA, November 28 - December 9, 2022}, 2022.
\newblock URL \url{http://papers.nips.cc/paper\_files/paper/2022/hash/82ad13ec01f9fe44c01cb91814fd7b8c-Abstract-Conference.html}.

\bibitem[Yao et~al.(2023)Yao, Zhao, Yu, Du, Shafran, Narasimhan, and Cao]{ReAct}
Shunyu Yao, Jeffrey Zhao, Dian Yu, Nan Du, Izhak Shafran, Karthik~R. Narasimhan, and Yuan Cao.
\newblock React: Synergizing reasoning and acting in language models.
\newblock In \emph{The Eleventh International Conference on Learning Representations, {ICLR} 2023, Kigali, Rwanda, May 1-5, 2023}. OpenReview.net, 2023.
\newblock URL \url{https://openreview.net/forum?id=WE\_vluYUL-X}.

\bibitem[Yin et~al.(2025)Yin, Wang, Pan, Lin, Wan, and Wang]{yin2025godelrsi}
Xunjian Yin, Xinyi Wang, Liangming Pan, Li~Lin, Xiaojun Wan, and William~Yang Wang.
\newblock G{\"o}del agent: A self-referential agent framework for recursively self-improvement.
\newblock In \emph{Proceedings of the 63rd Annual Meeting of the Association for Computational Linguistics (Volume 1: Long Papers)}, pp.\  27890--27913, 2025.

\bibitem[Zhang et~al.(2026{\natexlab{a}})Zhang, Wang, Zhou, Liao, Feng, Zhang, Wen, Li, Xiong, Qi, Tang, and Wen]{MemRL}
Shengtao Zhang, Jiaqian Wang, Ruiwen Zhou, Junwei Liao, Yuchen Feng, Weinan Zhang, Ying Wen, Zhiyu Li, Feiyu Xiong, Yutao Qi, Bo~Tang, and Muning Wen.
\newblock Memrl: Self-evolving agents via runtime reinforcement learning on episodic memory.
\newblock \emph{CoRR}, abs/2601.03192, 2026{\natexlab{a}}.
\newblock \doi{10.48550/ARXIV.2601.03192}.
\newblock URL \url{https://doi.org/10.48550/arXiv.2601.03192}.

\bibitem[Zhang et~al.(2026{\natexlab{b}})Zhang, Liu, Zhang, Hu, and Shao]{RetroAgent}
Xiaoying Zhang, Zichen Liu, Yipeng Zhang, Xia Hu, and Wenqi Shao.
\newblock Retroagent: From solving to evolving via retrospective dual intrinsic feedback.
\newblock \emph{CoRR}, abs/2603.08561, 2026{\natexlab{b}}.
\newblock \doi{10.48550/ARXIV.2603.08561}.
\newblock URL \url{https://doi.org/10.48550/arXiv.2603.08561}.

\bibitem[Zhao et~al.(2024)Zhao, Huang, Xu, Lin, Liu, and Huang]{ExpeL}
Andrew Zhao, Daniel Huang, Quentin Xu, Matthieu Lin, Yong{-}Jin Liu, and Gao Huang.
\newblock Expel: {LLM} agents are experiential learners.
\newblock In Michael~J. Wooldridge, Jennifer~G. Dy, and Sriraam Natarajan (eds.), \emph{Thirty-Eighth {AAAI} Conference on Artificial Intelligence, {AAAI} 2024, Thirty-Sixth Conference on Innovative Applications of Artificial Intelligence, {IAAI} 2024, Fourteenth Symposium on Educational Advances in Artificial Intelligence, {EAAI} 2024, February 20-27, 2024, Vancouver, Canada}, pp.\  19632--19642. {AAAI} Press, 2024.
\newblock \doi{10.1609/AAAI.V38I17.29936}.
\newblock URL \url{https://doi.org/10.1609/aaai.v38i17.29936}.

\bibitem[Zheng et~al.(2026)Zheng, Zhang, Ma, Yu, Zhu, Wu, Xu, Dong, Zhu, Huang, and Yu]{skillrouterskillroutingllm}
YanZhao Zheng, ZhenTao Zhang, Chao Ma, YuanQiang Yu, JiHuai Zhu, Yong Wu, Tianze Xu, Baohua Dong, Hangcheng Zhu, Ruohui Huang, and Gang Yu.
\newblock Skillrouter: Skill routing for llm agents at scale, 2026.
\newblock URL \url{https://arxiv.org/abs/2603.22455}.

\bibitem[Zhu et~al.(2023)Zhu, Chen, Tian, Tao, Su, Yang, Huang, Li, Lu, Wang, Qiao, Zhang, and Dai]{GhostIT}
Xizhou Zhu, Yuntao Chen, Hao Tian, Chenxin Tao, Weijie Su, Chenyu Yang, Gao Huang, Bin Li, Lewei Lu, Xiaogang Wang, Y.~Qiao, Zhaoxiang Zhang, and Jifeng Dai.
\newblock Ghost in the minecraft: Generally capable agents for open-world environments via large language models with text-based knowledge and memory.
\newblock \emph{ArXiv}, abs/2305.17144, 2023.
\newblock URL \url{https://api.semanticscholar.org/CorpusID:258959262}.

\end{thebibliography}
